\documentclass[conference]{IEEEtran}
\IEEEoverridecommandlockouts
\usepackage{cite}
\usepackage{amsmath,amssymb,amsfonts}
\usepackage{algorithmic}
\usepackage{graphicx}
\usepackage{textcomp}
\usepackage{xcolor}
\usepackage{subcaption}

\makeatletter
\newcommand{\newlineauthors}{%
  \end{@IEEEauthorhalign}\hfill\mbox{}\par
  \mbox{}\hfill\begin{@IEEEauthorhalign}
}
\makeatother

\newcommand{\IgnoredPart}[1]{\makebox[0pt][l]{#1}}

\def\BibTeX{{\rm B\kern-.05em{\sc i\kern-.025em b}\kern-.08em
    T\kern-.1667em\lower.7ex\hbox{E}\kern-.125emX}}

\usepackage{geometry}
\usepackage{afterpage}
\begin{document}
\newgeometry{
    top=72pt,     
    bottom=54pt,
    left=46pt,
    right=46pt
}
\afterpage{\restoregeometry}

\title{Hybrid Model Predictive Control with Physics-Informed Neural Network for Satellite Attitude Control\\
\thanks{
$^\dag$ Corresponding author}
\thanks{This work was supported by the the project PNRR-NGEU which has received funding from the MUR – DM 117/2023.
}}

\author{\IEEEauthorblockN{Carlo Cena\IgnoredPart{$^\dag$}}
\IEEEauthorblockA{\textit{Department of Electronics}\\
\textit{and Telecommunications,}\\
\textit{Politecnico di Torino}\\
Torino, Italy \\
carlo.cena@polito.it}
\and
\IEEEauthorblockN{Mauro Martini}
\IEEEauthorblockA{\textit{Department of Electronics}\\
\textit{and Telecommunications,}\\
\textit{Politecnico di Torino}\\
Torino, Italy \\
mauro.martini@polito.it}
\and
\IEEEauthorblockN{Marcello Chiaberge}
\IEEEauthorblockA{\textit{Department of Electronics}\\
\textit{and Telecommunications,}\\
\textit{Politecnico di Torino}\\
Torino, Italy \\
marcello.chiaberge@polito.it}
}
\maketitle

\begin{abstract}
Reliable spacecraft attitude control depends on accurate prediction of attitude dynamics, particularly when model-based strategies such as Model Predictive Control (MPC) are employed, where performance is limited by the quality of the internal system model. For spacecraft with complex dynamics, obtaining accurate physics-based models can be difficult, time-consuming, or computationally heavy. Learning-based system identification presents a compelling alternative; however, models trained exclusively on data frequently exhibit fragile stability properties and limited extrapolation capability. This work explores Physics-Informed Neural Networks (PINNs) for modeling spacecraft attitude dynamics and contrasts it with a conventional data-driven approach. A comprehensive dataset is generated using high-fidelity numerical simulations, and two learning methodologies are investigated: a purely data-driven pipeline and a physics-regularized approach that incorporates prior knowledge into the optimization process. The results indicate that embedding physical constraints during training leads to substantial improvements in predictive reliability, achieving a 68.17\% decrease in mean relative error relative. When deployed within an MPC architecture, the physics-informed models yield superior closed-loop tracking performance and improved robustness to uncertainty. Furthermore, a hybrid control formulation that merges the learned nonlinear dynamics with a nominal linear model enables consistent steady-state convergence and significantly faster response, reducing settling times by 61.52\%-76.42\% under measurement noise and reaction wheel friction.
\end{abstract}


\section{Introduction}
The effectiveness of a satellite attitude control system plays an important role in influencing pointing precision, operational reliability, and spacecraft longevity. Achieving these objectives is complicated by the nonlinear nature of spacecraft dynamics, persistent environmental perturbations, and actuator constraints. Reaction wheels (RWs), widely adopted for their smooth and accurate torque generation \cite{rw_ref}, introduce additional nonlinear effects, including friction-induced losses and saturation. Meanwhile, gravity-gradient forces, atmospheric drag, and geomagnetic disturbances continuously excite the rotational motion of the spacecraft, further complicating control design \cite{acs_challenges}. Historically, these challenges have been addressed through physics-based modeling and control methodologies \cite{IANNELLI2022401, generic_model_sat_att, manuel, java_riccati}, as well as, more recently, data-driven strategies \cite{react, rl_varying_masses, WU20241979, imit_learn_sat_att}. Analytical approaches provide interpretability and theoretical grounding but often rely on simplifying assumptions that limit their effectiveness under uncertainty and operational variability \cite{math11122614}. Conversely, machine learning models can represent highly complex behaviors directly from data but typically lack robustness guarantees, hindering their adoption in safety-critical aerospace applications \cite{davide}. These limitations motivate the development of hybrid modeling paradigms that combine physical knowledge with data-driven learning. Physics-Informed Neural Networks (PINNs) \cite{pinn1, pinn_loss_eletr, pinn_rizzo, pinn_sat_state_est} offer a promising solution by embedding governing equations into the optimization process, thereby promoting consistency with physical laws and improving extrapolation capabilities. This work targets attitude transition modeling conditioned on the system state and applied control torque, enabling seamless integration within predictive control architectures. We evaluate purely data-driven and physics-augmented learning strategies using high-fidelity datasets generated in the Basilisk simulation framework \cite{basilisk}. Using a Multilayer Perceptron (MLP) model, we show that the incorporation of physical constraints substantially improves both prediction accuracy and closed-loop behavior when embedded, alongside a linear model, in a hybrid Model Predictive Control (MPC) framework.
\begin{figure*}[!htb]
    \centering
    \centerline{\includegraphics[width=0.85\linewidth]{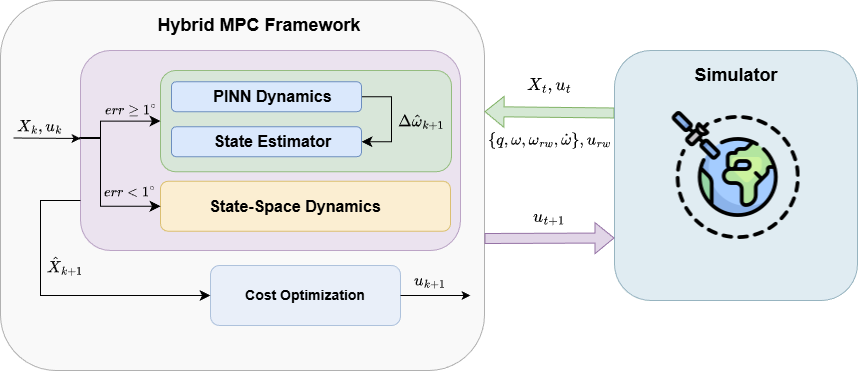}}
        \caption{Schematic of the proposed hybrid MPC with learned dynamics. When the attitude error is below 1 degree we switch from a nonlinear MPC with PINN state estimator (red) to a linear MPC (yellow).}
	\label{fig:MPC_scheme}
\end{figure*}
\subsection{Contribution}
The primary contributions of this work can be summarized as follows:
\begin{enumerate}
\item We develop a learning-based framework for modeling spacecraft attitude dynamics using a MLP architecture.
\item We design a physics-informed loss formulation that enhances robustness and generalization, with an automatic trade-off between empirical and physical objectives.
\item We conduct a systematic comparison between purely data-driven and physics-constrained learning, demonstrating a $68.17\%$ reduction in mean relative error over 10-step recursive prediction horizons.
\item We validate the control relevance of the learned models by embedding them into a hybrid PINN + linear MPC scheme, achieving improved closed-loop performance under realistic disturbances, including parameter uncertainties of up to $20\%$, observation noise of $3\%$, and reaction wheel friction.
\end{enumerate}

\subsection{Paper Organization}
Section~\ref{methodology} formalizes the satellite attitude dynamics problem, outlines the proposed model and its associated loss functions, and describes the MPC scheme in which the learned dynamics it embedded. Section~\ref{exp_setup} provides a detailed description of the dataset, the neural network architectures, the baseline MPC implementations, the experimental setup, and the evaluation metrics used to assess system performance. Section~\ref{results} reports and discusses the experimental findings. Finally, Section~\ref{conclusion} concludes the paper and highlights the main implications of our framework.

\section{Methodology}
\label{methodology}
\subsection{Satellite Attitude Dynamics}
\label{sec:dynamics_back}
A rigid spacecraft in Earth orbit is influenced by both actuator control torques and environmental disturbance torques, which may arise from gravity‑gradient effects, atmospheric drag, and geomagnetic interactions. Given the control torque in the body frame $N_{c}$, the spacecraft inertia matrix $I_{s}$, the reaction wheel inertia matrix $I_{rw}$, the satellite angular velocity $\omega$, the reaction wheel angular velocity $\omega_{rw}$, the total external torque $N_{e}$, and the skew‑symmetric operator $S(\omega)$ defined in~(\ref{eq:skewsym}), the reaction‑wheel‑actuated attitude dynamics of the spacecraft is expressed in~(\ref{eq:self_sup}).

\begin{equation}
    \label{eq:skewsym}
    S(\omega) = \begin{bmatrix} 
                        0 & -w_3 & w_2\\
                        w_3 & 0 & -w_1\\
                        -w_2 & w_1 & 0
                        \end{bmatrix}
\end{equation}

\begin{equation}
\label{eq:self_sup}
    \dot \omega = -I_{s}^{-1}S(\omega)I_{s}\omega - I_{s}^{-1}S(\omega)I_{rw}\omega_{rw}+I_{s}^{-1}N_{c}+I_{s}^{-1}N_{e}
\end{equation}

The dynamics of the reaction wheels angular velocity is given in eq. (\ref{eq:wrw}).

\begin{equation}
\label{eq:wrw}
\dot \omega_{rw} = I_{rw}^{-1}u_{rw} - \dot \omega
\end{equation}

The spacecraft and reaction wheel angular velocities are obtained by time‑integrating (\ref{eq:self_sup}) and (\ref{eq:wrw}). These quantities then permit the computation of the spacecraft attitude, represented by the quaternion $q$, through integration of (\ref{eq:quat}).

\begin{equation}
    \label{eq:quat}
    \dot q = \frac{1}{2} \Omega[\omega]q
\end{equation}

Where $\Omega[\omega]$ is given by eq. (\ref{eq:omegaw}).
\begin{equation}
    \label{eq:omegaw}
    \Omega[\omega] = \begin{bmatrix} 
                        0 & -\omega_0 & -\omega_1 & -\omega_2\\
                        \omega_0 & 0 & \omega_2 & -\omega_1\\
                        \omega_1 & -\omega_2 & 0 & \omega_0\\
                        \omega_2 & \omega_1 & -\omega_0 & 0
                        \end{bmatrix}
\end{equation}

\subsection{Problem Formulation}
Developing reliable attitude control remains difficult due to the spacecraft’s nonlinear dynamics and environmental uncertainties~\cite{acs_challenges}. Accurate modeling is therefore crucial, especially for model‑based controllers such as MPC and in the presence of state and parameter estimation errors. We propose a physics‑informed neural network (PINN) to approximate the spacecraft dynamics. A physics‑based loss term complements the data‑driven objective, improving generalization and robustness to noise. \figurename~\ref{fig:MPC_scheme} summarizes the approach. The PINN takes the spacecraft state—satellite and reaction‑wheel angular velocities—along with satellite angular acceleration and reaction‑wheel control torque, and predicts the change in satellite angular velocity for integration within a nonlinear MPC scheme. When integrated into the proposed hybrid MPC, the attitude dynamics are computed either using the PINN or a linear state-space formulation, depending on the current attitude error. Specifically, when the attitude error, \(err = 2arrcos(q_0)\) in \figurename~\ref{fig:MPC_scheme}, falls below 1 degree, the attitude dynamics estimator switches from the PINN model to the state-space model.

\subsection{Neural Network Architecture}
\label{sec:nn_architecture}
A MLP neural network is employed to learn the spacecraft attitude dynamics. The model input consists of the spacecraft inertia matrix concatenated with the current state, reducing dimensionality while retaining the quantities explicitly required in~\ref{eq:self_sup}. Omitting these terms would force the network to approximate them, increasing numerical error and diverting model capacity. The network outputs a vector of size $(3S, 1)$, representing the predicted trajectory over $S$ time steps with three variables per step. At time $t$, the models predict the angular‑velocity change $\Delta\hat{\omega}_{t+1}$ from the input state comprising the satellite angular velocity $\omega_{t}$, reaction‑wheel velocity $\omega_{rw,t}$, commanded torque $u_{rw,t}$, and the spacecraft angular acceleration $\dot{\omega}_{t}$, obtained via first‑order backward differencing.

\subsection{Physics-Informed Training}
\label{sec:pinn_training}
As previously discussed, we employed a combination of two loss terms. Given a batch size \textit{B}, the data-driven loss, \(L_{DD}\), defined in eq. (\ref{eq:data_driven}), corresponds to the Normalized Root Mean Squared Error (NRMSE) between the predicted change in angular velocity, \(\Delta \hat{\omega}\), and the ground-truth values from the dataset. The normalization is performed using the standard deviation of the ground-truth change in angular velocity, \(\sigma_{\Delta \omega}\), to scale the Root Mean Squared Error (RMSE).

\begin{equation}
    L_{\text{DD}} = \frac{\sqrt{\frac{1}{B}\sum_{i}^{B}(\Delta \hat \omega_{i} - \Delta \omega_{i})^2}}{\sigma_{\Delta \omega}}
\label{eq:data_driven}
\end{equation}

The physics-informed penalty term \(L_{PI}\) (eq. (\ref{eq:phys_inf_loss})) regularizes the training of the dynamics model by embedding the system’s physical laws directly into the loss function. Derived from the satellite attitude dynamics presented in Section \ref{sec:dynamics_back}, this term consists of a weighted sum of two components: (I) the NRMSE of the predicted change in angular acceleration, \(L_{\dot{\omega}}\), eq. (\ref{eq:dotw}), where \(\dot{\omega}\) is computed using eq. (\ref{eq:self_sup}); and (II) the MSE of the change in angular momentum, \(L_{h}\), eq. (\ref{eq:totangmom}).

\begin{equation}
\label{eq:phys_inf_loss}
L_{PI} = L_{\dot \omega} + pL_{h}
\end{equation}

The MSE term on the change in angular momentum is scaled by a factor \textit{p} set to $1e^{-2}$, selected through a grid search. To keep the physics-informed loss compact, the total external torque \(N_{e}\) in eq. (\ref{eq:self_sup}) is set to zero, since this loss acts primarily as a regularizer enforcing general dynamical consistency. External disturbances are already reflected in the training data and are therefore captured by the data-driven loss. This design choice keeps the physics-informed loss robust and independent of environment-specific assumptions, focusing instead on intrinsic system dynamics.
\begin{equation}
\label{eq:dotw}
L_{\dot \omega} = \frac{\sqrt{\frac{1}{B}\sum_{i}^{B}(\hat{\dot \omega} - \dot \omega)^2}}{\sigma_{\dot \omega}}
\end{equation}

\begin{align}
\label{eq:totangmom}
\begin{split}
L_{h}=& \frac{1}{B}\sum_{i}^{B}(||I_{s} \cdot (\omega+\Delta \hat \omega) + I_{rw} \cdot \hat \omega_{rw}|| \\
&- ||I_{s} \cdot (\omega+\Delta \omega) + I_{rw} \cdot \omega_{rw}||)^2
\end{split}
\end{align}

In equations (\ref{eq:dotw}) and (\ref{eq:totangmom}), \(\Delta \hat{\omega}\) and \(\hat{\dot{\omega}}\) denote the predicted change in angular velocity and the predicted angular acceleration, respectively, where \(\hat{\dot{\omega}} = \Delta \hat{\omega} / \Delta t\). \(\Delta t = 0.1\) is the controller timestep. The term \(\Delta \hat{\omega}_{rw}\) represents the change in RW angular velocity, computed using eq. (\ref{eq:wrw}) with \(\hat{\dot{\omega}}\). Equation (\ref{eq:total_loss}) summarizes the total loss, expressed as a weighted combination of the data-driven and physics-informed components controlled with the scalars \(\alpha\) and \(\beta\), where \(\alpha = (1 - \beta)\), and \(\beta\) is constrained to the interval \([0, 1]\).

\begin{equation}
    \label{eq:total_loss}
    L = \alpha L_{DD} + \beta L_{PI}
\end{equation}

A Lagrangian dual strategy \cite{lagr_paper}, which adaptively adjusts their importance based on model performance, is used.

\subsection{Non-linear MPC with Learned Dynamics} \label{mpc_cost_def}
The objective of this work is to learn a satellite’s attitude dynamics using a deep neural network, enabling more efficient and flexible spacecraft control. We evaluate the best learned models within a rest-to-rest maneuver scenario by integrating the learned dynamics into an MPC framework. This application also serves to assess how the learned model behaves when embedded in a model-based controller, particularly in terms of robustness to state-estimation errors and model uncertainties. The MPC state vector includes the quaternion \textit{q}, the spacecraft angular velocity $\omega$, the RWs angular velocity $\omega_{rw}$, and the angular acceleration $\dot{\omega}$, such that $x = \{q, \omega, \omega_{rw}, \dot{\omega}\}$. The PINN predicts the spacecraft and RW dynamics from the current state $x_{t}$ and RW torque input $u_{t}$, and these predictions are then integrated to compute the next state $x_{t+1}$, as expressed in equations~\ref{eq:dw_mpc}–\ref{eq:dotq_mpc}.

\begin{align}
    \label{eq:dw_mpc}
    &\Delta \hat{\omega}_{t+1} = PINN_{\theta}(\omega_t, \omega_{rw, t}, \dot \omega_t, u_t,\theta) \\
    \label{eq:dotw_mpc}
    &\hat{\dot{\omega}}_{t+1} = \frac{\Delta \hat{\omega}_{t+1}}{\Delta t} \\
    \label{eq:dotwrw_mpc}
    &\hat{\dot{\omega}}_{rw,t+1} = I_{rw}^{-1} u_{t} - \hat{\dot{\omega}}_{t+1} \\
    \label{eq:dotq_mpc}
    &\hat{\dot{q}}_{t+1} = \frac{1}{2} \Omega[\omega_{t}]q
\end{align}

In particular, the PINN is used to estimate the angular‑velocity increment $\Delta \hat{\omega}_{t+1}$, which is then used to compute the angular acceleration $\hat{\dot{\omega}}_{t+1}$ (eq.~\ref{eq:dotw_mpc}) and the RWs accelerations $\hat{\dot{\omega}}_{rw}$ by substituting the estimated acceleration into eq.~(\ref{eq:wrw}). During inference only the first predicted state is provided to the MPC as the estimated next state. The MPC's cost function is defined in eq. (\ref{eq:mpc_cost_f}).
\begin{equation}
\label{eq:mpc_cost_f}
C = \sum_{k=0}^{n-1}\left( x_{k}^{T}Qx_{k} + u_{k}^TCu_{k}
    + \Delta u_k^T R \Delta u_k\right)
    + x_{n}^{T}Qx_{n}
\end{equation}
Where $u_k$ is an abbreviation of the commanded RWs torque $u_{rw}$ at time step \textit{k}, and $\Delta u_{k}$ is the difference between the RWs torque required at the previous time step and the current one. \textit{C} and \textit{R} are diagonal cost matrices used to reduce the torques $u_{rw}$ and to limit their variation, smoothing their trajectories. Their non-zero elements are equal to $1e-1$, while $Q^{13 \times 13} =$ \text{diag}(10000, 10000, 10000, 10000, $1e^{-2}, 1e^{-2}, 1e^{-2}, 1e^{-4}, 1e^{-4}, 1e^{-4}, 1e^{-2}, 1e^{-2}, 1e^{-2}$). Finally, \textit{n} is the horizon of the MPC, which in our experiments is set to $n = 10$.

\section{Experimental setups}
\label{exp_setup}
\subsection{Dataset}
To train and evaluate the models, we generated a dataset using the Basilisk simulator \cite{basilisk}. The simulation considers a cubesat in Low Earth Orbit actuated solely by RWs and executing attitude maneuvers under an MRP feedback control module available in Basilisk. In addition to the commanded torques, the satellite is subjected to several environmental disturbances, including gravity–gradient effects, magnetic disturbances, and atmospheric drag. The relevant satellite parameters are summarized in Table~\ref{tab:datasets_info}.
\begin{table}[htb]
    \caption{We show the satellite inertia matrix \(I_S\), the RWs inertia matrix \(I_{rw}\), the satellite mass, the maximum torque acting on the RWs \(u_{rw}\), the maximum RWs speed \(\omega_{rw}\).}
    \centering
    \begin{tabular}{@{}ccccc@{}}
        \hline
        \(I_{s}\) & \(I_{rw}\) & \begin{tabular}{@{}c@{}}Mass\\\ [Kg]\end{tabular} & \begin{tabular}{@{}c@{}}Max \(u_{rw}\)\\\ [Nm]\end{tabular} & \begin{tabular}{@{}c@{}}Max \(\omega_{rw}\)\\\ [rpm]\end{tabular}\\
        \hline\hline
        \begin{tabular}{@{}c@{}}5.700, 0.045, 0.002\\0.045, 3.300, 0.012\\0.002, 0.012, 6.100\end{tabular} & \begin{tabular}{@{}c@{}}0.001, 0., 0.\\0., 0.001, 0.\\0., 0., 0.001\end{tabular} & 58 & 0.05 & 6000\\
        \hline
    \end{tabular}
    \label{tab:datasets_info}
\end{table}
We conducted 300 simulations by randomly varying the initial attitude and orbital position to generate the main dataset, and an additional 50 simulations with modified satellite mass and inertia matrix to assess model robustness to approximately 10\% parameter variations along all axes—an important consideration since these quantities are not known with perfect accuracy and may vary during operations. In each run, the satellite was initialized with a random reaction–wheel angular velocity uniformly sampled from the range [-300, 300] rpm. Each simulation lasted 3 minutes with a control and sampling period of 0.1 seconds, and an integration time step of 0.001 seconds using a fourth-order Runge–Kutta method. The dataset was constructed by forming input–output pairs as follows: each input $x_{i}$ has shape (1, 12) and is defined as $x_i = \{\omega, \omega_{rw}, u_{rw}, \dot{\omega}\}$, while the target $y_i$ corresponds to the change in angular velocity, i.e., $y_i = \Delta \omega_{i}$. The resulting dataset was randomly split into training and validation subsets using a 67–33\% ratio.

\subsection{Models}
In our experiments, we evaluated multiple models with different inputs and architectures. Our goals were to achieve high performance in estimating the next attitude state from the current state and the commanded torque, while also ensuring a low inference time. Beyond varying the number of FC layers and the number of units per layer, we also assessed whether including the satellite and RWs inertia information could improve performance. To do so, we concatenated this information to the model input and again before the final layer. The most effective configuration was obtained by concatenating \(I_{s}\) and \(I_{rw}\) directly to the model input. After identifying the best-performing model structures, we conducted additional tests on the number of predicted steps, \(S\). The model was trained to output all future \(S\) steps in a single forward pass, with the two losses computed by iteratively propagating the state while keeping the torques constant. The best performance was achieved with \(S=10\). During inference, only the first predicted state was used for evaluation or as the estimated state within the MPC. The best-performing model consisted of 4 FC layers with 16 units each. This configuration was then used to analyze the impact of concatenating the satellite and RWs inertia parameters from the training set, \(I_{s}\) and \(I_{rw}\), to the input state. The results confirmed that including these inertia parameters yielded improved performance.

\subsection{Traditional MPC} \label{sec:trad_mpc}
In this work, we compare our framework against two MPC baselines: one based on a state–space formulation (linear MPC) and another relying on a non-linear formulation. Using the MPC setup introduced in Section \ref{mpc_cost_def}, we apply the same input state and cost matrices together with the non-linear dynamics described in Section \ref{sec:dynamics_back}, and modify them only when adopting the state–space model. Specifically, we define the input state as \(x = \{q, w\}\), set all diagonal terms of the matrix \(Q^{6 \times 6}\) to 1000, and leave all other parameters unchanged. The discrete state–space model was implemented following \cite{manuel}.

\subsection{Evaluation Metrics \& Experiments} \label{sec:eval_metrics_exp}
To evaluate our models, we employed a set of metrics designed to assess their performance both as standalone regressors and as components within MPC controllers. These two perspectives correspond to the two evaluation stages carried out in our study. First, we trained several neural networks with different architectures and hyper-parameters to identify the most effective model for estimating the next attitude state from the current one and the commanded RWs torque, whose results are shown in Section \ref{subsec:dyn_learn_ex}. The metrics used were the Mean Relative Error (MRE) and the physics error, i.e., the physics-informed loss. Results were averaged across all time steps and simulations. For both metrics, we report the single-step prediction performance as well as the performance when predicting the next 10 steps under the assumption of constant command torque. Second, in Section \ref{subsec:mpc_control}, we used the best-performing neural networks to evaluate a range of non-linear and linear MPCs. For each model, we conducted 300 Monte Carlo simulations to analyze robustness to noise. In these simulations, we added a Gaussian-distributed random error of up to 3\% to each state variable, a continuously distributed random error of up to 10\% to the inertia matrix, and up to 20\% to the satellite mass. Additionally, RWs friction was modeled with linear dynamics up to $50\%\pm12.5\%$ of the maximum wheel velocity. The initial attitude was randomly sampled within the interval \((\frac{\pi}{8}, \frac{\pi}{2})\). It is important to note that the neural networks were trained on noise-free simulations without RWs friction. To evaluate these experiments, we present the average steady-state error over the final minute of simulation, the time required for the attitude error to fall below 1 degree (Settling Time), and the distribution of the trajectories.

\subsection{Implementation Details}
\label{Impl_det}
The experiments were performed on a computer with an Intel i7-9700K 3.60GHz CPU, 64 GB of RAM, and a GeForce RTX 2080 Ti. The models and loss functions have been implemented in torch and we used the "do-mpc" library to implement the MPC with AI as dynamics estimator. The models have been trained on the GPU using a batch size of 16384 and tested on CPU.

\section{Results}
\label{results}
\subsection{Attitude Dynamics Regressor}
\label{subsec:dyn_learn_ex}
We compared the performance of model trained solely with the data-driven loss against the physics-informed neural network trained with both losses. Table \ref{tab:dynamic_learn_exp0} reports the outcomes of these experiments. Specifically, the table shows for each configuration the average MRE and physics error, both for single-step prediction and for the 10-step self-loop predictions, where the model’s output is recursively fed back as input. MLP-DD corresponds to the model trained only with the data-driven loss, while MLP-LD employed the Lagrangian dual method to automatically adjust the weight between data-driven and physics-informed loss.
\begin{table}
    \caption{Results obtained when evaluating the MLP models as regressors. MLP-DD uses only the data-driven loss, while MLP-LD uses a Lagrangian dual approach.}
    \centering
    \begin{tabular}{@{}ccccc@{}}
         \hline
         \textbf{Experiment} & \textbf{MRE} & \begin{tabular}{@{}c@{}}\textbf{Physics} \\ \textbf{Error}\end{tabular} & \begin{tabular}{@{}c@{}}\textbf{MRE} \\ \textbf{self-loop}\end{tabular} & \begin{tabular}{@{}c@{}}\textbf{Physics Error} \\ \textbf{self-loop}\end{tabular} \\
         \hline\hline
         MLP-DD & 19.67 & 24.38 & 22.63 & 137.34 \\
         MLP-LD & \textbf{6.26} & \textbf{17.18} & \textbf{6.96} & \textbf{95.86} \\
         \hline
    \end{tabular}
    \label{tab:dynamic_learn_exp0}
\end{table}
The experiments showed that the model trained with both losses achieved superior performance, improving both the physics error and the MRE. To further validate these findings and assess their statistical significance, we computed p-values using the Wilcoxon test. Following standard practice, results with p-values above 0.05 are considered not significant. In our case, all p-values were below this threshold, confirming that the observed improvements are statistically significant.

\subsection{MPC Controllers}
\label{subsec:mpc_control}
\begin{figure*}[ht!]
    \centering
    \begin{subfigure}[t]{0.32\textwidth}
        \centering
        \includegraphics[width=\columnwidth]{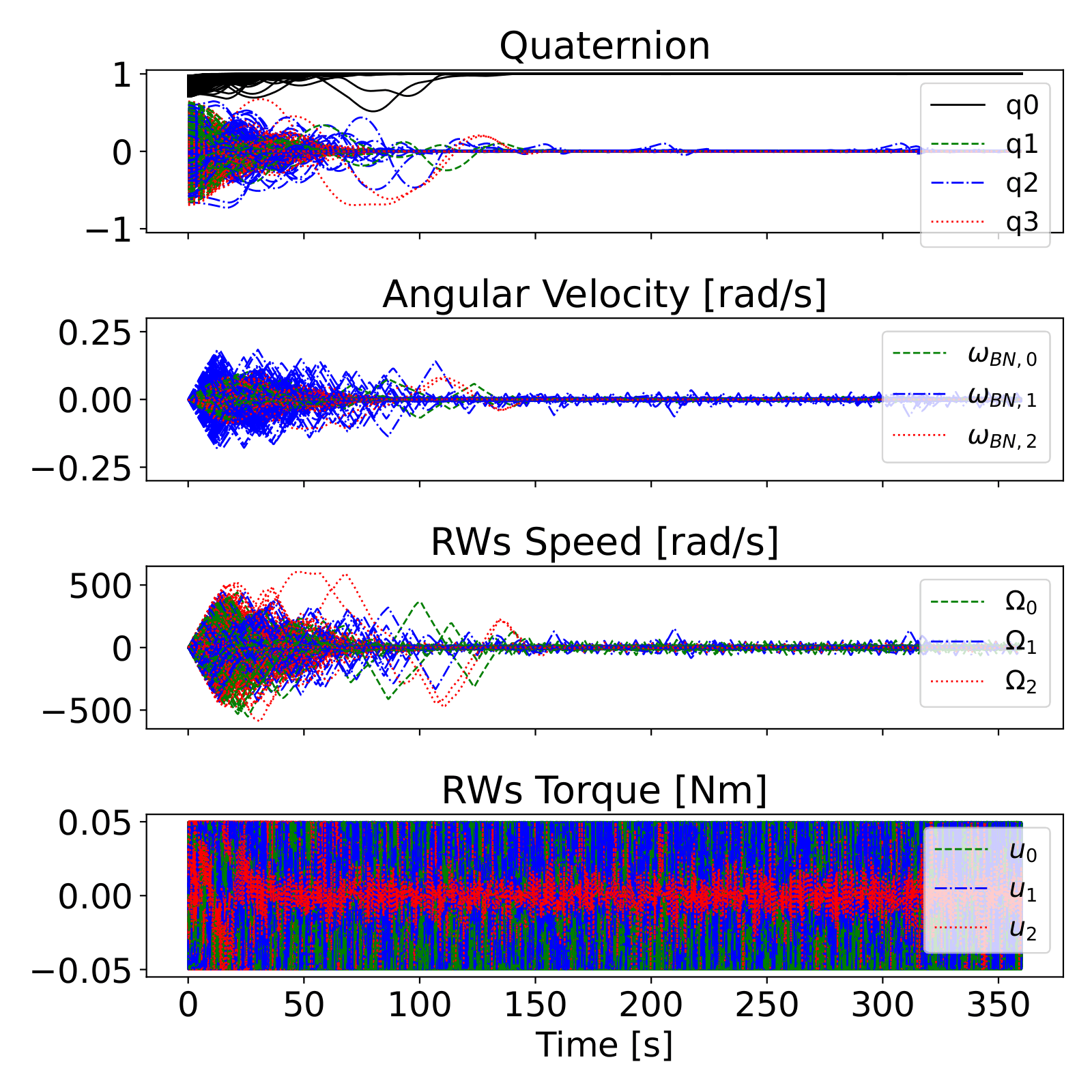}
        \captionsetup{font=footnotesize, justification=centering}
        \caption{MLP-LD}
    \end{subfigure}
    \begin{subfigure}[t]{0.32\textwidth}
        \centering
        \includegraphics[width=\columnwidth]{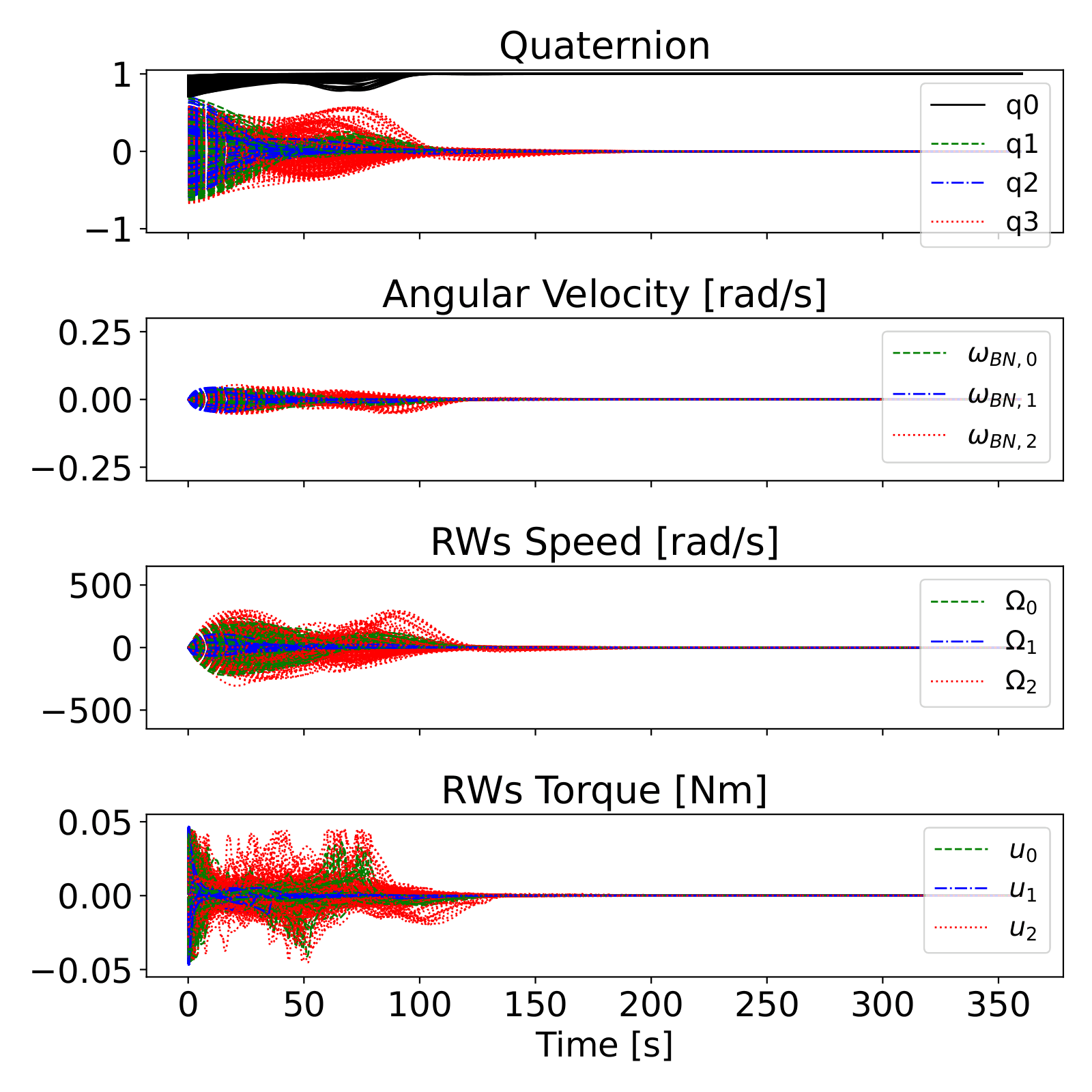}
        \captionsetup{font=footnotesize, justification=centering}
        \caption{Linear MPC}
    \end{subfigure}
    \begin{subfigure}[t]{0.32\textwidth}
        \centering
        \includegraphics[width=\columnwidth]{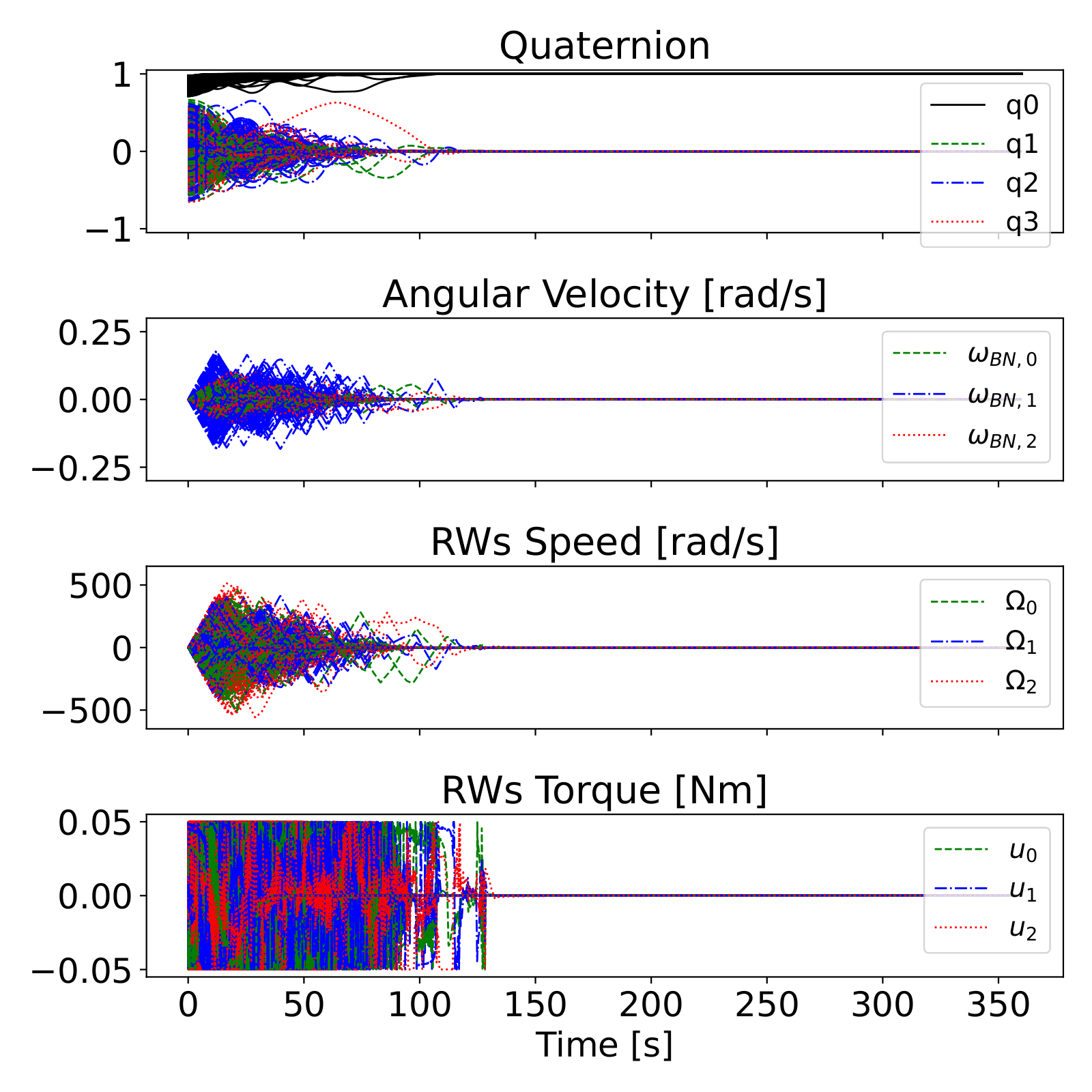}
        \captionsetup{font=footnotesize, justification=centering}
        \caption{MLP-LD + Linear}
    \end{subfigure}
    \caption{300 MC simulation with parameter estimation errors, state estimation noise and RWs friction for MLP-LD (left), traditional MPC with linear (center) and MLP-LD + Linear (right).}
    \label{fig:traj_MPC_noise_est}
\end{figure*}

Given the results presented above, we selected the MLP-LD model as the dynamics predictor within the MPC controller to compare the performance of our best model against the two traditional MPC controllers introduced in Section \ref{sec:trad_mpc}. In \figurename~\ref{fig:traj_MPC_noise_est}, we show the trajectories from 300 Monte Carlo simulations performing rest-to-rest maneuvers over 360 seconds using the PINN-based controller, the state-space MPC and the proposed hybrid MPC framework. Across these simulations, we randomly varied the orbital position and the initial satellite attitude, and introduced parameter estimation errors of up to 20\% in the satellite mass and up to 10\% in the inertia matrix. Additionally, state estimation was corrupted with up to 3\% Gaussian noise, and the reaction wheels were subject to friction as described in Section \ref{sec:eval_metrics_exp}.
\begin{table}[!htb]
    \caption{Results of the experiments performed to study the robustness-to-noise of the PINN model when used in a MPC with several sources of noise. We show the median over 300 MC simulations.}
    \centering
    \begin{tabular}{@{}ccc@{}}
         \hline
         \textbf{Experiment} & \begin{tabular}{@{}c@{}}\textbf{Steady-State Error} \\ \textbf{[degrees]}\end{tabular} & \textbf{Settling Time [s]} \\
         \hline\hline
         MLP-LD & 0.2901 & 47.40\\
         MLP-LD + Linear & \textbf{0.0023} & \textbf{43.90}\\
         Non-Linear MPC & 0.1215 & 186.2\\
         Linear MPC & 0.0024 & 114.1\\
         \hline
    \end{tabular}
    \label{tab:noise_res}
\end{table}
The results indicate that all controllers can reach the desired attitude with a steady-state error below 1 degree, though the use of the MLP-LD neural network as dynamics estimator for the complete maneuver leads to a higher final error and continuous high RWs torques. The MPC using the hybrid MPC scheme (MLP-LD + Linear) instead achieves the fastest convergence despite the presence of noise, outperforming the other controllers in terms of settling time, and achieving comparable results to the linear MPC in terms of steady-state error. However, the RWs torque distributions show that this model leads to considerably higher torques than necessary. Table \ref{tab:noise_res} reports the median settling time, defined as the time required to reach a pointing error below 1 degree, and the median steady-state error over the final minute of each trajectory, i.e., the angular difference in degrees between the current and target quaternion. The results show that the state-space MPC requires significantly more time, 114.1 seconds compared to the 43.9 seconds needed by MLP-LD + Linear model, to achieve convergence.

\section{CONCLUSIONS}\label{conclusion}
In this study, we introduce a framework for learning spacecraft attitude dynamics using a multilayer perceptron (MLP) neural network trained with a loss function that combines data-driven supervision with a physics-informed penalty, where the relative weight is dynamically adjusted during training through the Lagrangian dual approach. Our analysis indicates that incorporating physical knowledge into the loss function significantly enhances model performance, resulting in notable reductions in the Mean Relative Error (MRE) for 10-step predictions. The robustness of our approach is demonstrated through 300 Monte Carlo simulations with a high-fidelity simulator, which show resilient closed-loop behavior and resistance to observational noise and reaction wheel friction. We evaluated the average steady-state errors and settling times for both purely data-driven and physics-informed models, comparing them against traditional MPC controllers. Our results indicate improvements in settling time of up to 76.42\%. We believe that the proposed approach exhibits strong generalization capabilities and could be extended to more complex non-linear scenarios, such as satellite berthing or docking with non-cooperative targets, or the control of under-actuated spacecraft. These situations involve greater modeling and control challenges, such as discontinuous dynamics, partial observability, and restricted actuation, which align well with the strengths of hybrid data-driven and physics-informed learning.

\section*{Acknowledgment}
This work has been developed with the contribution of the Politecnico di Torino Interdepartmental Centre for Service Robotics (PIC4SeR) (https:// pic4ser.polito.it) and Argotec (https://argotecgroup.com). Computational resources were provided by HPC@POLITO (http://hpc.polito.it). The cover in \figurename~\ref{fig:MPC_scheme} has been designed using resources from Flaticon.com.

\bibliographystyle{IEEEtran}
\bibliography{refs}

\end{document}